\documentclass[letterpaper, 10 pt, conference]{ieeeconf}  

\IEEEoverridecommandlockouts                              

\usepackage{amsmath} 
\usepackage{amssymb}  
\usepackage{siunitx}
\usepackage{nicefrac}
\usepackage{paralist}
\usepackage{soul}
\usepackage{graphicx}

\newcommand{\norm}[1]{\left\lVert#1\right\rVert}
\usepackage{pgfplots,lipsum}
\usepackage{booktabs}
\pgfplotsset{compat=newest}
\usepackage{scalefnt}
\usepackage{hyperref}
\usepackage{multirow}
\usepackage{subcaption}
\usepackage{wrapfig}
\usepackage[super]{nth}
\usepackage{fancyhdr}

\title{\LARGE \bf
Efficiency and Equity are Both Essential:
A 
Generalized 
Traffic Signal Controller with Deep Reinforcement Learning}

\author{Shengchao Yan, Jingwei Zhang, Daniel B\"uscher, Wolfram Burgard
\thanks{This project was funded through the Priority Programme “Cooperative Interacting Automobiles” of the German Science Foundation DFG.}
\thanks{
All authors are with the Department of Computer Science, University of Freiburg, Germany. Wolfram Burgard is also with the Toyota Research Institute, Los Altos, USA.
        {\tt\small \{yan, zhang, buescher, burgard\} @cs.uni-freiburg.de}}%
}

\begin{document}

\maketitle
\thispagestyle{fancy}

\begin{abstract}
Traffic signal controllers play an essential role in today's traffic system. However, the majority of them currently is not sufficiently flexible or adaptive to generate optimal traffic schedules.
In this paper we present an approach to learn policies for signal controllers using deep reinforcement learning aiming for optimized traffic flow.
Our method uses a novel formulation of the reward function that simultaneously considers efficiency and equity. 
We furthermore present a general approach to find the bound for the proposed \textit{equity factor}
and
we introduce the \textit{adaptive discounting} approach 
that greatly stabilizes learning and helps to maintain a high flexibility of green light duration. 
The experimental evaluations on both simulated and real-world data demonstrate that our proposed algorithm 
achieves state-of-the-art performance (previously held by traditional non-learning methods) 
on a wide range of traffic situations.
\end{abstract}

\section{Introduction}
\label{sec:intro}

Traffic congestion is enormously expensive in terms of fuel and time 
and many cities all over the world suffer from it~\cite{scorecard}.
Moreover, the emissions of road transport have been considered as the main cause for air pollution~\cite{silva2016impact,denney2012national}.
To alleviate traffic congestion and the associated problems,
smarter and cleaner vehicles have been investigated~\cite{lipson2016driverless,menendez17iv}.
However, an alternative way to improve the effectivity of road traffic is to optimize the scheduling of traffic lights.

In this paper,
we focus on reducing congestion by improving automated traffic light controllers. 
More specifically, 
we focus on traffic signal controllers (TSCs) for isolated intersections~\cite{papageorgiou2003review}, i.e.,
signalized intersections
whose traffic is unaffected by any other controllers or supervisory devices.

The performance of conventional fixed-time or actuated TSCs is limited by the restricted setup and the relative primitive sensor information available. Recently, adaptive TSCs~\cite{li2013survey} attracted more attention due to their high degree of flexibility.
Advances in perception and vehicle-to-everything (V2X) communication~\cite{guler2014using} could make such controllers even better by providing additional  (real-time) information,
such as  locations and velocities of the vehicles.
With more detailed information available,
adaptive TSCs have the potential to provide optimal control according to current traffic situations.
One approach to achieve this is to consider traffic signal optimization as a scheduling problem~\cite{li2013survey,xie2012schedule},
in which a junction is considered as a production line and the input vehicles as different products to be processed.
However,
this type of methods suffers from the curse of dimensionality which limits their applicability to small numbers of vehicles~\cite{abdulhai2003reinforcement}.
As a result,
these methods in general only satisfy real-time requirements for either oversimplified intersections or under small traffic flow rates.

A recent line of research proposes to design adaptive TSCs based on deep reinforcement learning (DRL).
DRL has been shown to reach state-of-the-art performance in various domains~\cite{mnih2015human,schulman2017proximal}. 
However, 
we believe that the performance of DRL approaches in the traffic domain can be pushed further, 
in particular with regards to the following limitations:
\begin{itemize}
    \item
        Most previous approaches have focused on improving efficiency, which is calculated according to the throughput of intersections. However, we argue that the equity of the travel time of individual vehicles is also of vital importance.
        Previous works have been mostly evaluating in scenarios with relatively low traffic flow,
        in which case the trade-off between efficiency and equity might not have a great influence on the performance of the controller.
        However, in dense traffic with nearly- or even over-saturated intersections and unbalanced traffic density on incoming lanes,
        the efficiency-equity trade-off can be an important factor.
    \item
        The flexibility of adaptive TSCs has not been sufficiently explored. 
        Instead, 
        most approaches employ fixed green traffic light duration or fixed traffic light cycles.
    \item
        Previously proposed DRL agents are trained and evaluated in relatively simplified traffic scenarios: 
        very few traffic demand episodes with limited variation or
        evenly distributed flow for each incoming lane~\cite{genders2019open}.
        Thus, their experimental results might not be sufficient indicators of their performance in real traffic scenarios.
    \item
        Current DRL-based approaches have shown performance improvement mainly against fixed-time or actuated TSCs. 
        They either have not compared with state-of-the-art adaptive TSCs, such as the Max-pressure controller~\cite{varaiya2013max}, or do not surpass state-of-the-art performance~\cite{genders2019open,genders2016using}. 
\end{itemize}

To overcome these limitations,
we present a novel method that introduces the following innovations:
\begin{itemize}
    \item
        An \textit{equity factor} to trade off efficiency (average travel time) against equity (variance of individual travel times) as well as a solution to calculate a rough bound for it.
    \item
        An \textit{adaptive discounting} method to account for the issues brought by transitional phases of traffic signals,   
        which is shown to substantially stabilize learning.
    \item
        A learning strategy that surpasses state-of-the-art baselines. 
        It is generic with regards to different traffic flow rates, 
        traffic distributions among incoming lanes and intersection topologies.
\end{itemize}

In line with the aforementioned DRL approaches,
we conduct experimental studies in the traffic simulation environment SUMO~\cite{SUMO2018}.
We show that our method achieves state-of-the-art performance, 
which had been held by traditional non-learning methods, 
on a wide range of traffic flow rates with varying traffic distributions on the incoming lanes. 

\section{Related Works}
\label{sec:related}

In traditional fixed-time TSC designs~\cite{papageorgiou2003review},
the traffic flow rates at intersections are treated as constants,
and the green-red phases for each route are scheduled in a cyclic manner.
Then the duration for each green phase is optimized using history flow rates.
The Uniform TSC with the same fixed duration for all green phases and the Webster's method~\cite{Webster1958} with pre-timed duration according to latest traffic history are usually used as baselines in TSC works~\cite{genders2019open}.
As the real traffic flow rates generally vary across lanes and across time,
the performance of such TSCs could be very restricted.

Actuated TSCs~\cite{papageorgiou2003review} make use of loop detectors,
which are electromagnetic sensors mounted within the road pavement.
Such sensors can detect the incoming vehicles and estimate their velocity when they pass by,
so that actuated TSCs can dynamically react to the vehicles driving into the intersection.
Yet,
their performance are still restricted due to the limited information provided by the sensor.

Since decades researchers have investigated on developing adaptive TSCs, 
which can schedule traffic lights acyclic and with flexible green phase duration according to the real-time traffic situation. 
Some early works like~\cite{gartner1983opac,lowrie1990scats} have been largely applied in real traffic designs.
Yet it is still believed that the performance of TSC can be further improved. 
In recent years, 
analytical~\cite{guler2014using,yang2016isolated}, 
heuristic~\cite{varaiya2013max,gregoire2014capacity} 
and learning-based~\cite{genders2019open,genders2016using,el2014design,wei2018intellilight} approaches have been proposed.
Among these, the heuristic Max-pressure method~\cite{varaiya2013max} is reported to be holding state-of-the-art performance~\cite{genders2019open}.
DRL-based methods hold great promise with the possibility to learn generalized and flexible controller policies by interacting with traffic simulators, 
and that they could provide scheduling decisions in real-time, 
as opposed to some non-learning methods that need optimization iterations before giving out each decision.

A few works have deployed DRL for isolated intersection TSCs~\cite{genders2019open,genders2016using,VanDerPol16LICMAS,liang2019deep}. 
However, 
none of them were able to surpass state-of-the-art performance achieved by the Max-pressure method. 
Each of these method proposes its own reward functions for training the agent, 
but the connection between them has not been clear.  
In this work we attempt to give such an analysis of those different reward functions that have been proposed (Sec.~\ref{sec:reward-types}). 

While efficiency has been the main objective for most of these works, 
some previous algorithms actually had considered equity implicitly. 
They~\cite{wei2018intellilight,VanDerPol16LICMAS,wei2019survey} design the reward as a weighted sum of several different quantities about the intersection. 
However, 
finding the optimal weighting is non-trivial. 
In this paper we instead propose an \textit{equity factor} along with a method to calculate its rough bound. 

\section{Methods}
\label{sec:methods}

\subsection{Background}
We consider the task of TSC in standard reinforcement learning settings.
At each step,
from its state $s\in\mathcal{S}$ the agent selects an action $a\in\mathcal{A}$ according to the policy $\pi(\cdot|s)$.
It then transits to the next state $s'\in\mathcal{S}$ and receives a scalar reward $r\in\mathbb{R}$.
The state and action spaces and the reward function in our work are discussed in the next subsections.

For learning the optimal policy that maximizes the discounted (by $\gamma$) cumulative expected rewards,
we use proximal policy optimization (PPO)~\cite{schulman2017proximal} as the backbone DRL algorithm.
For a policy $\pi_{\theta}$ parameterized by $\theta$,
PPO maximizes the following objective:
\begin{align}
    \mathcal{J}_{\theta}=\mathbb{E}_t
    \Big[
        &\min
            \Big(
                \rho_t(\theta) A_t,
                \text{clip}\left( \rho_t(\theta), 1-\epsilon, 1+\epsilon \right) A_t
            \Big)
        \nonumber\\&+ \beta_\text{entropy} \cdot H \Big( \pi_{\theta}(s_t) \Big)
    \Big],
\end{align}
where the expectation is taken over samples collected by following $\pi_{\theta_\text{old}}$,
and $\rho_t({\theta})=\nicefrac{\pi_{\theta}(a_t|s_t)}{\pi_{\theta_\text{old}}(a_t|s_t)}$ is the importance sampling ratio.
$H$ represents the entropy of the current policy,
$\beta_\text{entropy}$ adjusts the strength of entropy regularization.
$A_t$ is a truncated version (on trajectory segments of length up to $K$) of the generalized advantage estimator~\cite{Schulmanetal_ICLR2016},
which is an exponentially-weighted average (controlled by $\lambda$):
\begin{align}
    A_t = \delta_t + (\gamma\lambda) \delta_{t+1} + \dots + (\gamma\lambda)^{K-1-t}\delta_{K-1},
\end{align}
where $\delta_t = r_t + \gamma V_{\phi_\text{old}}(s_{t+1}) - V_{\phi_\text{old}}(s_t)$.
The value function $V_{\phi}$,
parameterized by $\phi$,
is learned by minimizing the following loss (with coefficient $\beta_\text{value}$):
\begin{align}
    \mathcal{L}_{\phi} = \beta_\text{value} \cdot \mathbb{E}_t
    \left[
        \norm{V_{\phi}(s_t) - \Big(V_{\phi_\text{old}}(s_t) + A_t\Big)}^2_2
    \right].
\end{align}

\subsection{Action Space}
\label{sec:action}
\begin{figure}[t]
    \centering
    \begin{subfigure}[b]{4.8cm}
        \centering
        \includegraphics[width=4.5cm]{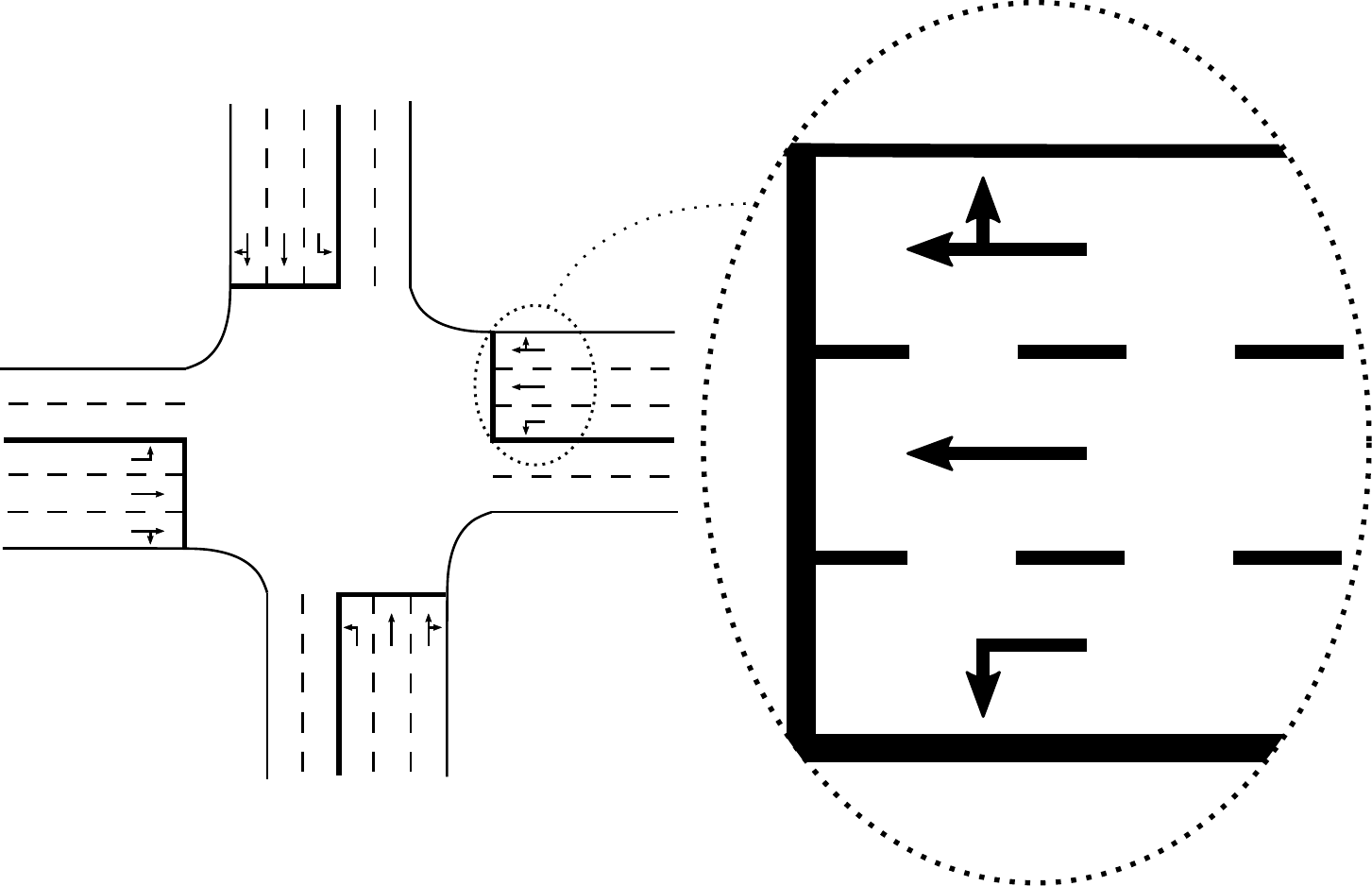}
        \caption{Four-road intersection with three incoming lanes for each road.}
        \label{fig:intersection}
    \end{subfigure}\hfill
    \begin{subfigure}[b]{2.8cm}
        \centering
        \includegraphics[width=2.5cm]{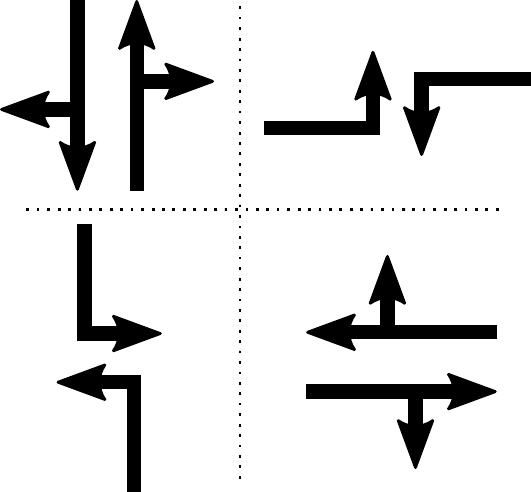}
        \caption{Four green phases (actions) for the intersection. 
        }
        \label{fig:actions}
    \end{subfigure}
    \caption{The intersection and its corresponding action space.}
    \label{fig:intersection_action}
\end{figure}

We carry out our method on a four-road intersection where each road contains three incoming lanes 
(one forward-only, one forward+right-turning, one left-turning, Fig.~\ref{fig:intersection}). 
We note that our approach can easily generalize to other intersections by adjusting the state and action representations accordingly.

The agent has an action space of size $4$:
while one of the two sets of facing directions (north and south, east and west) has only red light, the other set can schedule either of the following two traffic light signal combinations (Fig.~\ref{fig:actions}):
\begin{itemize}
 \item Green for the forward-only and forward+right-turning lanes and red for the rest;
\item Green for the left-turning lanes and red for the rest.
\end{itemize}
In order to give the agent more flexibility,
we set the duration for each of the $4$ actions as $1$ second.

We note that choosing one action means scheduling a distinct green phase.
During the transition between different green phases,
yellow or all-red phases must be scheduled.
In our work,
a $3\si{\second}$-yellow and $2\si{\second}$-all-red phase is scheduled before activating a new green phase.
We denote the constant $\mathsf{T}_{yr}=5\si{\second}$ as the duration for the yellow-red phase.

Due to this setting,
if two different actions (green phases) are scheduled consecutively,
the effective duration of the second action is $6\si{\second}$ instead of $1\si{\second}$;
while if the same action (green phase) is scheduled twice in a row,
then the effective duration for the second action is still $1\si{\second}$.
During the learning process,
the aforementioned two scenarios should not be treated equally.
To cope with this we propose the method of \textit{adaptive discounting} which will be presented when discussing the reward function (Sec.~\ref{sec:reward}).

\subsection{State Space}
\label{sec:state}
At each process step,
the state $s_t$ the agent receives is comprised of the following components:
 \begin{itemize}
    \item
        The distance along the lane to the traffic light and the velocity of each vehicle that has not passed the light and is within $150\si{\meter}$ range (each lane has a maximum capacity of $19$ vehicles) to the center of the intersection.
        A block of $19\times2$ scalars in the state vector is reserved for the vehicles in each incoming lane. 
        The vehicles' states in each block are sorted according to their distance values.
        The order of lanes in the state vector has to be kept unchanged.
        All the values are normalized to be within $[-1,1]$.
        If any lane does not reach its maximum capacity,
        the corresponding position and velocity values will be set to $1$ and $-1$.
    \item
        The action of the last step $a_{t-1}$ (in one hot encoding so a $4$-dimensional vector).
    \item
        A counter that contains for each action the time in seconds since its last execution.
        The $4$-dimensional vector is normalized by $500\si{\second}$.
        This component along with the last action $a_{t-1}$ helps to avoid state-aliasing.
\end{itemize}

\subsection{Reward Function}
\label{sec:reward}
Several different reward functions have been proposed in previous works to train DRL agents for controlling traffic signals.
However,
the reasoning behind different designs have not been clearly presented,
also the connections between those different choices and the different effects they are causing have not been thoroughly analyzed.
We attempt for such an analysis below,
which indicates that the vanilla versions of those rewards tend to result in policies that only consider time efficiency (average travel time in an intersection).
We then propose solutions that also take equity (variance of individual travel time) into consideration.

\begin{figure*}[t]
    \center
    \includegraphics[width=1.0\linewidth]{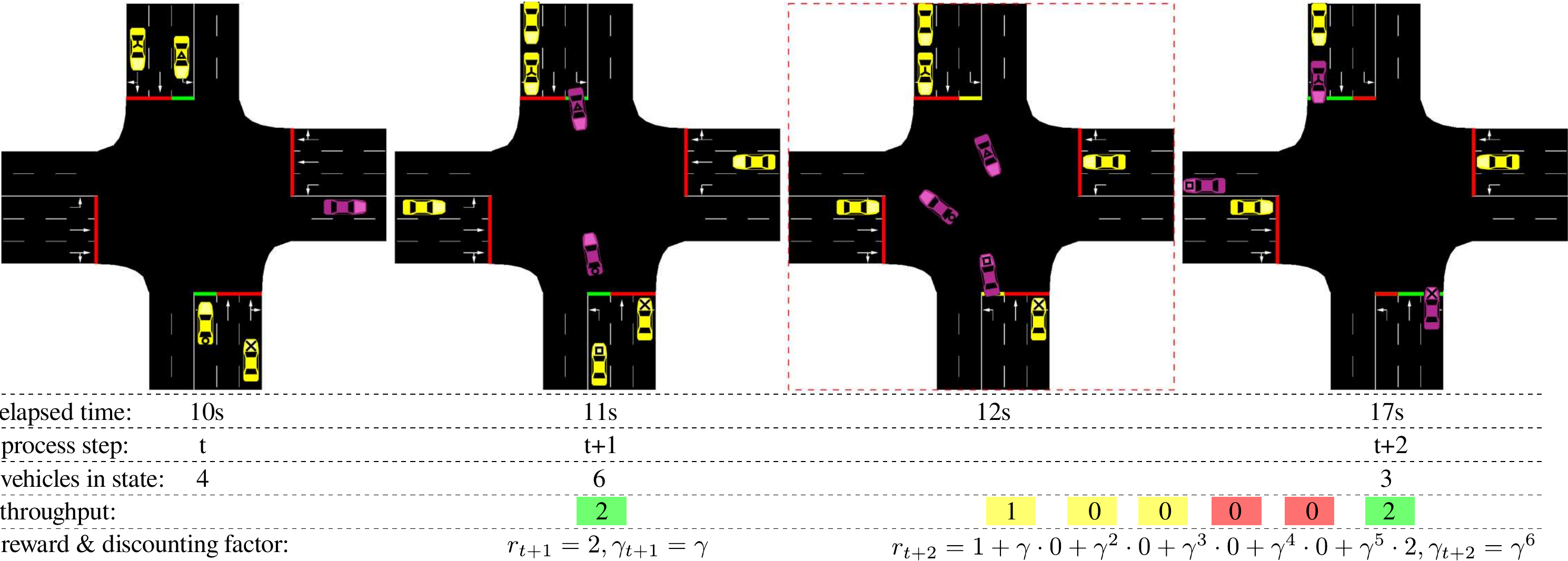}\\
    \centering
    \caption{
        Illustration on the proposed \textit{adaptive discounting} (Sec.~\ref{sec:reward-adaptive}), 
        as well as several important concepts in the traffic intersection domain (Sec.~\ref{sec:reward-defs}). 
        In the figure we show each released car with a distinguish symbol on top.  
        As for the colors, 
        the cars in yellow are those that have not yet passed through the traffic light,  
        and they would be depicted in purple immediately after they pass through their traffic lights 
        (judged by the head of the car). 
        The cars in yellow are considered in the state representation, 
        while the cars that turned from yellow to purple are calculated into the throughput. 
        The \nth{1}, \nth{2} and \nth{4} sub-figures correspond respectively to system elapsed time: $\{10,11,17\}\si{\second}$, 
        and to learning process step: $\{t,t+1,t+2\}$. 
        Since the action $a_{t+1}$ chooses to schedule a different green phase than that of $a_t$, 
        a $3\si{\second}$-yellow and $2\si{\second}$-red phase will be scheduled before the new green phase. 
        The \nth{3} sub-figure in red dashed bounding box shows the \nth{1} second in the yellow phase. 
        In previous works, 
        the discounting has been conducted with respect to the process step. 
        While we propose to discount according to system elapsed time which is shown in our experiments to be of vital importance for stable learning.}
    \label{fig:state-seq}
\end{figure*}

\subsubsection{Definitions}
\label{sec:reward-defs}
We first give the definitions of several important concepts in traffic intersection systems.
We visualize the important ones in Fig.~\ref{fig:state-seq}.
\begin{itemize}
    \item Total number of vehicles in the intersection ($N$):
        At $t$,
        the number of vehicles in the intersection system $N_t$ is the total number of vehicles that are within a certain range to the intersection center (e.g. $150\si{\meter}$) but have not yet passed through the corresponding traffic lights.
    \item Throughput ($N^\text{TP}$):
        The number of vehicles that pass through the traffic lights of their corresponding incoming lanes within $\left(t-1,t\right]$ is denoted $N^{\text{TP}}_t$.
    \item Travel time ($T_\text{travel}$): 
        For a single vehicle,
        its travel time is counted as the time period starting from when it enters the intersection and ending when it passes through the traffic light.
        The total travel time of the intersection is the summation of the individual $T_\text{travel}$ of each vehicle in the intersection.
        We note an equivalent way of calculating the total travel time is to count $N_t$ at every second 
        and sum it over a given time period.
    \item Delay time ($T_\text{delay}$):
        Similar to travel time,
        except that a constant is subtracted from each individual travel time:
        $T_\text{delay}=T_\text{travel}-\mathsf{T}_\text{free}$,
        where $\mathsf{T}_\text{free}$ is the constant time length for a vehicle to pass through the intersection system with no cars ahead and green lights always on.
    \item Traffic flow rate ($F$):
        The number of vehicles that pass through an intersection in unit time. 
        A commonly used unit is the number of vehicles per hour $\nicefrac{\text{v}}{\si{\hour}}$.
    \item Saturation flow rate ($\mathsf{F}_\text{s}$):
        This is a constant representing the traffic flow rate for one lane under the condition that 
        the traffic light stays green during unit time and that the flow of traffic is as dense as it could be~\cite{Bester07saturationflow}.
\end{itemize}

\subsubsection{Reward Function Categories}
\label{sec:reward-types}
Given the above definitions,
the majority of the reward functions proposed in the TSC domain can be categorized into the following two types:
\begin{itemize}
    \item Throughput-based reward functions $\mathcal{R}^\text{TP}$~\cite{wei2018intellilight}.
        The vanilla form of this type
        uses the throughput $N_{t}^{\text{TP}}$ as the reward for step $t$. 
        Learning on this reward function means maximizing the cumulative throughput of the intersection.
        The change in throughput $N^\text{TP}_t - N^\text{TP}_{t-1}$ has also been used as a reward function~\cite{salkham2008collaborative}.
    \item Travel-time-based reward functions $\mathcal{R}^\text{TT}$~\cite{genders2019open,genders2016using,el2014design,wei2018intellilight,VanDerPol16LICMAS}.
        As mentioned before,
        the total travel time of an intersection for a given period of time $[\tau_\text{start}, \tau_\text{end}]$ can be calculated as the summation of $N_t$ during that time: $\sum_{\tau_\text{start}}^{\tau_\text{end}}N_t$.
        The vanilla reward function of this type thus uses $-N_t$ as the reward for step $t$.
        We note that $N_t=N_{t-1}-N^\text{TP}_t+N^\text{in}_t$ where $N^\text{in}_t$ denotes the number of new vehicles input into the system from $t-1$ to $t$, 
        which is commonly assumed to be determined solely by the traffic flow distribution thus is out of the control of TSC. 
        Learning on this reward function would result in policies that minimize cumulative travel time.
        Reward functions utilizing the change of cumulative delay time between actions and 
        the total delay time of the intersection 
        have also been investigated.
\end{itemize}

The above description indicates that maximizing cumulative throughput and minimizing total travel time 
could both result in policies that puts efficiency in the top priority. 
During research we observed that throughput-based reward generally leads to more stable learning with smaller variance across different runs. 
Therefore, we focus on throughput-based reward in the following content.

\subsubsection{Adaptive Discounting}
\label{sec:reward-adaptive}
When calculating rewards of either of the two reward categories, 
the two scenarios discussed in Sec.~\ref{sec:action} 
should not be discounted in the same way. 
We propose the method of \textit{adaptive discounting} that properly discount for those scenarios and is shown to be critical for convergence in our experiments. 

We illustrate this method under the throughput based reward $R^{\text{TP}}_t=N^{\text{TP}}_t$ in Fig.~\ref{fig:state-seq}: 
At system elapsed time $10\si{\second}$ the reinforcement learning process is at step $t$. 
The action $a_t$ is chosen that schedules green lights for the left-turning lanes for the north-south roads. 
Transitioning from $t$ to $t+1$, 
the throughput reward obtained is $r_{t+1}=2$. 
This is a normal RL iteration and no special adjustments need to be done.
But at step $t+1$ when the system elapsed time is at $11\si{\second}$, 
the action $a_{t+1}$ is chosen to schedule green lights for the forward+right turning directions of the north-south roads, 
which is a different green phase than that of $a_{t}$.
This means a $3\si{\second}$-yellow and a $2\si{\second}$-all-red phase will be automatically scheduled before the new green phase. 
The $5\si{\second}$ intermediate phase and the chosen $1\si{\second}$ green phase are both within step $t+2$ of the learning process. 
During this step the throughput obtained at elapsed times $\{12,13,14,15,16,17\}\si{\second}$ are $\{1,0,0,0,0,2\}$. 
With no special treatment when calculating the reward for step $t+2$ it would be $r_{t+2}=3$. 
But this could lead to undesired properties since the agent gets the intermediate phase "for free" for collecting extra rewards whenever it chooses to schedule a different green phase, 
and that the subsequent states are not sufficiently discounted. 
Furthermore, 
given that the throughput of two episodes matches at every system elapsed second, 
the agent should obtain exactly the same return, even with different traffic light schedules. 
However, with the transitional phases it is not anymore a one-to-one mapping between the system time and the process step.   
So when discounting according to process steps, 
those two episodes of interest could lead to different returns. 
This issue has been overlooked in the current literature of DRL based TSC designs~\cite{genders2019open, el2014design}. 
Thus we propose the method of \textit{adaptive discounting} 
to account for the mismatch between the two timing paradigms, 
in which we discount the reward according to system elapsed time instead of learning process steps. As a result, the reward for $t+2$ is calculated as: 
\begin{align}
    r_{t+2}=
        1+
        \gamma\cdot0+
        \gamma^2\cdot0+
        \gamma^3\cdot0+
        \gamma^4\cdot0+
        \gamma^5\cdot2, 
\end{align}
and a discount factor of $\gamma^6$ instead of $\gamma$ will be used for the subsequent reward or value.

\subsubsection{The Equity Factor}
\label{sec:reward-equity}

Having presented the \textit{adaptive discounting} technique,  
now we present the \textit{equity factor} for reward functions for training TSC. 
The aforementioned two types of reward functions 
(throughput-based and travel-time based) 
both treat efficiency,
i.e. average travel time of the intersection as the major concern.
Equity,
the variance of individual travel times,
is not explicitly considered.
Take the following scenario as an example:
Assuming that the north-south roads are saturated,
while the east-west roads have lighter traffic,
the policy to maximize the cumulative throughput should always keep the north-south traffic lights green,
while keeping the east-west lights red.
Consequently,
the vehicles on the east-west roads might have to wait for an intolerable long time to pass through the intersection.
This is due to that in the vanilla reward definitions (Sec.~\ref{sec:reward-types}),
every vehicle contributes equally to the throughput or to the travel time,
regardless of how long it has been waiting.

Following the above analysis,
we propose to use the vehicle's travel time together with an equity factor $\eta$ in the reward function.
The basic idea is to adapt the contribution of each vehicle to the throughput-based reward according to its travel time in the intersection while passing the traffic light.
Instead of just counting value $1$ when a vehicle passes through, we consider three ways to incorporate $\eta$ into the reward calculation:
linear ($\eta \cdot T_\text{travel}$),
power (${T_\text{travel}}^\eta$) and base ($\eta^{T_\text{travel}}$).
Since simply scaling the rewards does not change the value function landscape,
we mainly considered the power and base forms. 
During research our experiment results show that the power form equity factor leads to convergence to better policies than the base form. 
Therefore, we focus on the analysis of the ${T_\text{travel}}^\eta$ in the following.

To define the proper range of $\eta$, 
two special scenarios are considered.
\begin{itemize}
    \item Scenario 1:
        Only one vehicle is before the traffic light,
        and its travel time at step $t$ is $\tau$.
        With the equity factor $\eta$ and the discount factor $\gamma$,
        the return contributed by this vehicle would be $\tau^\eta$ if it passes through the traffic light at $t$,
        and $\gamma\cdot(\tau+1)^\eta$ if one second later.
        We require $\tau^\eta>\gamma\cdot(\tau+1)^\eta$ so that releasing this vehicle sooner is more desired.
        With this we get $\eta<\nicefrac{\ln(\gamma)}{\ln{\frac{\tau}{\tau+1}}}$,
    \item Scenario 2:
        One lane with green light is over-saturated,
        while a single car is waiting at red light in another lane.
        In the case where the over-saturated lane always has green light on and the single vehicle is never released,
        the highest return for any state is:
        \begin{align*}
        G^\text{e}
        ={\mathsf{T}_\text{free}}^\eta \left( 1 + \gamma^{\frac{1}{\mathsf{F}_\text{s}}} + (\gamma^{\frac{1}{\mathsf{F}_\text{s}}})^2 + \cdots \right)= \nicefrac{{\mathsf{T}_\text{free}}^\eta} {1-\gamma^{\frac{1}{\mathsf{F}_\text{s}}}}
        \end{align*}
        (denoted as $G^\text{e}$ as in this case \textbf{e}fficiency is the top priority).
        If the waiting vehicle is released at step $t$ when its travel time is $\tau$, 
        the upper limit of the return the system can obtain at state $s_t$ is:
        \begin{align*}
            \sup(G^\text{e+e}) =
                &\mathsf{T}_\text{free}^\eta +
                      \tau^\eta \cdot \gamma^{\mathsf{T}_\text{yr}} +
                \nonumber\\&
                      \nicefrac{
                        \big(\mathsf{T}_\text{free}+2\cdot\mathsf{T}_\text{yr}+1\big)
                        ^\eta \cdot \gamma^{2\cdot\mathsf{T}_\text{yr}+1}
                      }
                      {1-\gamma^{\frac{1}{\mathsf{F}_s}}}
        \end{align*}
        (we use $G^\text{e+e}$ since this strategy cares about \textbf{e}fficiency and \textbf{e}quity).
        The three terms in the summation are all calculated out of the best case scenario
        (the traffic light on the saturated lane turns yellow then red for a total of $\mathsf{T}_\text{yr}$ elapsed time,
        then the light on the single vehicle lane turns green for one second then turns yellow)
        to get the upper limit:
        the first term is the reward obtained from the vehicle on the saturated lane that manages to pass through
        at the beginning of the yellow phase;
        the second term is contributed by the single vehicle passing through the traffic light in its $1\si{\second}$ green phase;
        the last term is the summation of the reward obtained by the vehicles on the saturated lane after the green phase switches back to this lane.
        We require $G^\text{e}<\sup(G^\text{e+e})$ to release the single vehicle after certain travel time $\tau$.
\end{itemize}
With these analysis a range of $\eta$ can be found. We note that this is a rough calculation under our system settings as for example
the traffic flow in the saturated lane does not recover instantaneously to $\mathsf{F}_s$ after the green light switches back.
Nevertheless the analysis gives a general solution to calculate a rough bound for $\eta$.
The experimental results show that the desirable TSC policies could be learned in this bound.


\section{Experiments}
\label{sec:exp}

\subsection{Experimental Setup}
\label{sec:exp-setup}

We conduct experiments using the urban traffic simulator SUMO~\cite{SUMO2018} 
and evaluate the trained agents in both simulated one-hour traffic demand episodes (with the intersection type described above) and a real-world whole-day traffic demand (with a different type of intersection in Freiburg, Germany). 
Both intersections have a speed limit of $50\nicefrac{\si{\kilo}\si{\meter}}{\si{\hour}}$.
We compare with the following common baselines in the TSC domain:
\begin{itemize}
	\item Uniform: 
		This controller circulates ordered green phases in the intersection.
		Each green phase is scheduled for a same fixed period,
		the duration of which is a hyper-parameter of this algorithm.
	\item Webster's~\cite{Webster1958}:
		Same as the Uniform controller,
		it schedules traffic phases in a cyclic manner.
		But 
		each phase duration is adjusted in accordance with the latest traffic flow history.
		It has three hyperparameters:
		the length $T_\text{history}$ of how long the traffic flow history to take into account for deciding the phase duration for the next $T_\text{history}$ period,
		and the minimum and maximum duration for one complete cycle.
	\item Max-pressure~\cite{varaiya2013max}:
		Regarding vehicles in lanes as substances in pipes,
		this algorithm favors control schedules that maximizes the release of pressure between incoming and outgoing lanes.
		More specifically,
		with incoming lanes containing all lanes with green traffic light in a certain phase,
		and outgoing lanes being those lanes where the traffic from the incoming lanes exit the intersection system,
		this controller tends to minimize the difference in the number of vehicles between the incoming and outgoing lanes.
		The minimum green phase duration is a hyper-parameter.
\end{itemize}
We note that previous learning methods were not able to surpass the state-of-the-art performance held by the non-learning method Max-pressure TSC~\cite{genders2019open}.

Regarding our network architecture for the intersection in Fig.~\ref{fig:intersection},
the input size for both the policy network $\theta$ and the value network $\phi$ is
$4+4+2\cdot19\cdot12=464$.
Then $\theta$ consists of fully connected layers of sizes
$2\,048$ (ReLU),
$1\,024$ (ReLU)
and $4$ (SoftMax),
where $4$ is the size of the action space.
For $\phi$ the fully connected layers are of sizes
$2\,048$ (ReLU),
$1\,024$ (ReLU)
and $1$.
We perform a grid search to find the hyperparameters.
We use $2.5\mathrm{e}{-5}$ as the learning rate for the Adam optimizer,
$1\mathrm{e}{-3}$ as the coefficient for weight decay.
For PPO,
we use $32$ actors,
$0.2$ for the clipping $\epsilon$.
In each learning step a total number of around $20$ mini-batches of size $1\,000$ is learned for $8$ epochs.

\subsection{Training}
Previous methods focused on relatively limited traffic situations, 
for example 
a single one-hour demand episode~\cite{el2014design} and traffic input less than $3\,000\nicefrac{\text{v}}{\si{\hour}}$~\cite{genders2019open,genders2016using}.
In this paper we challenge our method to experience a wider range of traffic demand. 
For the four-way junction we consider, 
the upper bound of the traffic flow can be calculated as $4\cdot\mathsf{F}_s$, 
where $\mathsf{F}_s$ is the saturation flow rate for one incoming lane.  
This maximum flow is reached when all $4$ forward-going lanes of either the north-south or the east-west roads have green lights and are in full capacity.  
However, 
this extreme scenario rarely happens in real traffic. 
In our experiments we found that the intersection already starts to saturate with around $3\,000\nicefrac{\text{v}}{\si{\hour}}$ of total traffic input. 
In our training we set the range of traffic flow rate to be $\left[F_{\min}, F_{\max}\right]=\left[0,6\,000\nicefrac{\text{v}}{\si{\hour}}\right]$ 
which is much wider than that used in previous works. 

With this flow rate range, 
we sample traffic demand episodes for training.  
Each episode is $1\,200\si{\second}$ long and defined by these randomly sampled parameters:
the total traffic flow at the beginning and end $F_\text{begin}$ and $F_\text{end}$, 
and for each incoming lane its traffic flow ratio of the total input at the beginning and end. 
$F_\text{begin}$ is randomly sampled from $\left[F_{\min}, F_{\max}\right]$.
Then $F_\text{end}$ is sampled uniformly within 
$[\max(F_{\min}, F_{\text{begin}}-1\,500), 
  \min(F_{\max}, F_{\text{begin}}+1\,500)]$.
The flow ratios are decided by sampling $12$ uniform random numbers then normalized by their sum. 
The traffic flow during the episode is then linearly interpolated.  
The sampled episodes with possibly big change of traffic flow and unbalanced distribution should be enough to cover real traffic scenarios.

\subsection{Evaluation during Training}

\begin{figure}[t]
    \centering
    \includegraphics[width=1.0\linewidth]{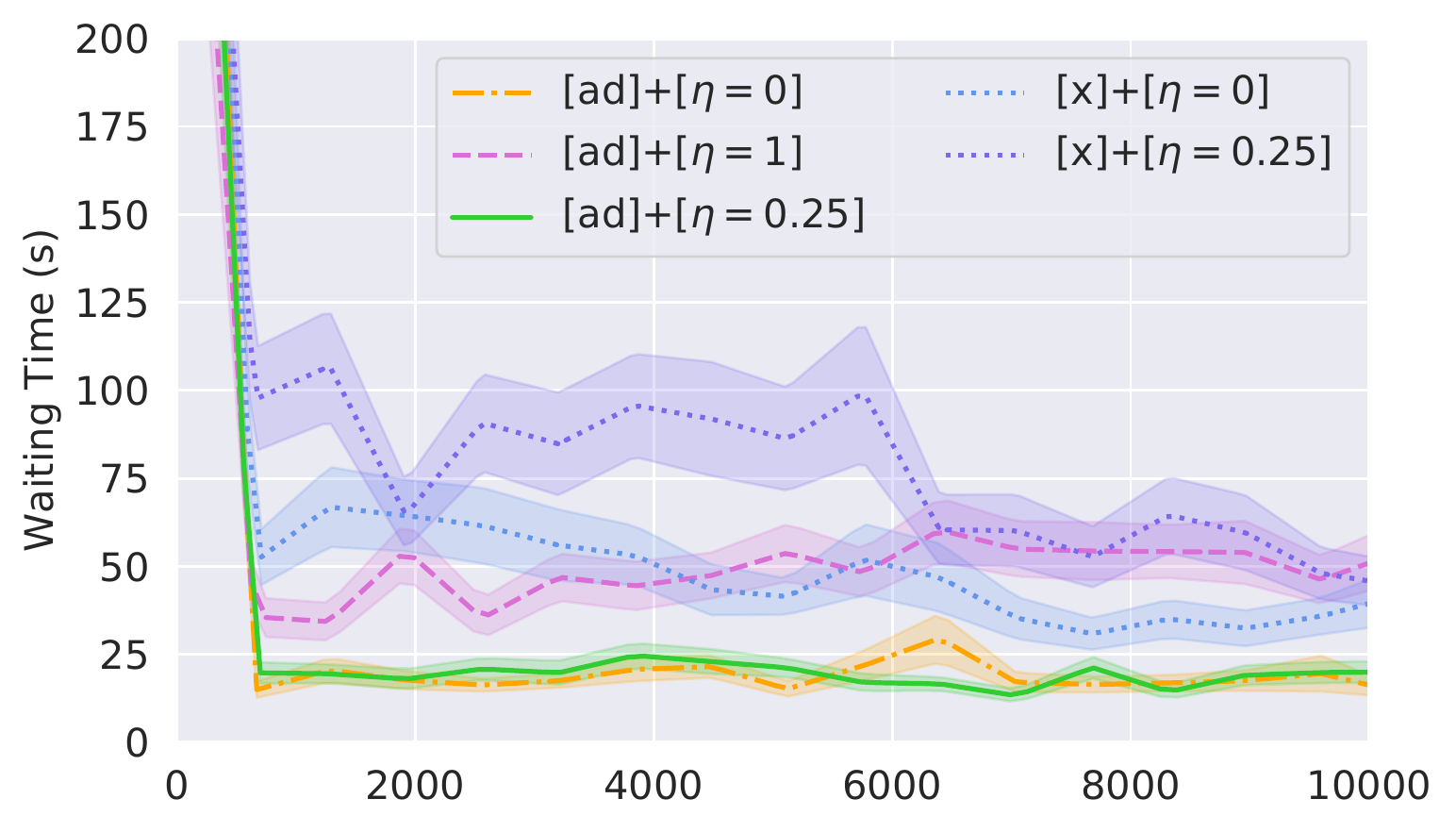}
    \includegraphics[width=1.0\linewidth]{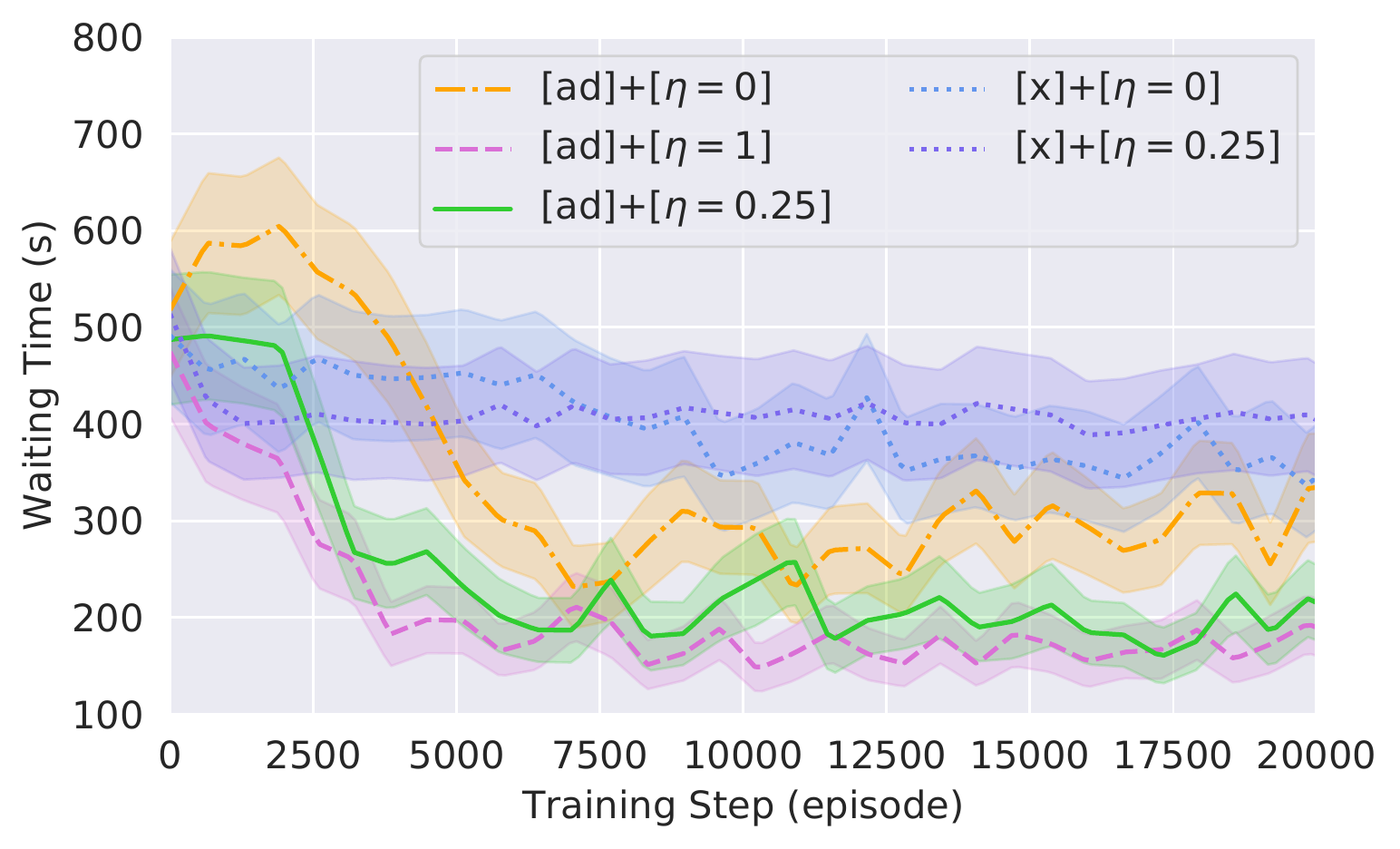}
    \caption{
    Waiting time obtained in evaluation during training for all agent configurations under ablation study. 
    Each plot shows the mean with $\pm\nicefrac{1}{5}$ standard deviation over 3 non-tuned random seeds 
    (we show $\nicefrac{1}{5}$ of the standard deviation for clearer visualization).
    The upper figure shows the logs of the evaluators of traffic flow range $[500,1\,500]$, 
    while the lower one shows that of $[4\,500,5\,500]$. 
    The vehicles passed the traffic light are not considered for the waiting time. The waiting time for a vehicle is calculated with $T_\text{episode}-T_\text{in}$, where $T_\text{episode}$ is the episode duration and $T_\text{in}$ is the time when it enters the intersection. 
    }
    \label{fig:ablation}
\end{figure}

During training, 
we conduct evaluation to monitor the learning progress every $20$ learning steps,  
which corresponds to $640$ episodes experienced by $32$ actors. 
For each evaluation phase $5$ evaluators are deployed, 
corresponding to traffic flow ranges of 
$\left[500,1\,500\right]$,
$\left[1\,500,2\,500\right]$,
$\left[2\,500,3\,500\right]$,
$\left[3\,500,4\,500\right]$ and 
$\left[4\,500,5\,500\right]$ respectively. 
Each evaluator samples traffic demand for evaluation in the corresponding range 
similar to how training episodes are sampled except that the flow rates at the beginning and end are independently sampled from the same corresponding range.

An ablation study is conducted to analyze the individual contributions of different components in our proposed algorithm. 
The plots are shown in Fig.~\ref{fig:ablation}, where the following agent configurations are compared: $[\times]+[\eta=0]$, 
$[\times]+[\eta=0.25]$,
$[\text{ad}]+[\eta=0]$, 
$[\text{ad}]+[\eta=1]$, 
$[\text{ad}]+[\eta=0.25]$. 
$[\text{ad}]$ means the agent utilizes \textit{adaptive discounting} while $[\times]$ means not; 
$[\eta=\cdot]$ denotes the value of the equity factor used by the agent, 
where the $[\eta=0]$ agents, which use exactly the vanilla throughput-based reward, care only about efficiency while the $[\eta=1]$ ones favor equity. 

Interestingly, 
from Fig.~\ref{fig:ablation} 
we can observe that the two agents without the technique of \textit{adaptive discounting} struggle to learn successful policies in both low and high flow rates. 
We can also observe the influence of the equity factor $\eta$:  
the $[\text{ad}][\eta=0]$ agent who does not care about equity converges to a better policy than the $[\text{ad}][\eta=1]$ agent in lower traffic density, 
while the latter agent outperforms the former one in denser traffic. 
This makes sense, 
since with little traffic input 
the equity problem is not critical, 
while with higher traffic flow the intersection could be saturated with continuously growing queues even under optimal policies. 
The efficiency-first policies favor releasing more vehicles in saturated traffic, thus vehicles in other lanes could have long waiting time. 

We observe that the $[\text{ad}]+[\eta=0.25]$ configuration obtains the best performance across different traffic flow rates, 
thus this is used for the agent \textit{Ours} in the following experiments. 

Having compared the plots of travel time (for released vehicles) and waiting time (for not released vehicles), we notice that the average waiting time always decreases during training when the policy gets better, while the average travel time may vary in different ways. This is because the travel time only considers the released vehicles. Some initial poor policies may choose the same action all the time, which leads to fast throughput for vehicles on the lanes with green light while extremely long waiting time for other vehicles. 
The waiting time, however, considers only the vehicles not passed the intersection during the episode. As the policy gets better, the number of vehicles staying in the intersection at the end becomes smaller. In order to show the training process clearly, we choose to use the plot of waiting time.

\subsection{Evaluation on Simulated Traffic Demand}

\begin{figure}[t]
	\centering
	\setlength{\fboxrule}{0pt}
	\fbox{
		\scalefont{0.8}
		\begin{tikzpicture}[
		every axis/.style={
			ybar stacked, axis on top,
			height=6cm, width=\columnwidth,
			bar width=0.3cm,
			ymajorgrids, tick align=inside,
			yminorgrids, tick align=inside,
			grid style={dashed,gray,opacity=0.4},
			enlarge y limits={value=.1,upper},
			ymin=0, ymax=300,
			ytick={0,100,200,300},
			minor ytick={50,150,250},
			axis x line*=bottom,
			axis y line*=left,
			tickwidth=0pt,
			subtickwidth=0pt,
			enlarge x limits=true,
			legend style={
				at={(0.5,1.)},
				anchor=north,
				legend columns=-1,
				/tikz/every even column/.append style={column sep=0.5cm}
			},
			ylabel={Mean \& STD of Travel Time (s)},
			xlabel={Range of Total Traffic Input (v/h)},
			symbolic x coords={
				500-1500,1500-2500,2500-3500,3500-4500,
				4500-5500},
			xtick=data,
		}
		]
		
		\begin{axis}[bar shift=-0.45cm, hide axis]
		\addplot [draw=none, fill=black] coordinates
		{(500-1500, 51.9) (1500-2500, 67.3) (2500-3500, 99.8) (3500-4500, 146.8) (4500-5500, 166.4)};
		\addplot [draw=none, fill=black, fill opacity=0.4] coordinates
		{(500-1500, 13.6) (1500-2500, 26.0) (2500-3500, 54.8) (3500-4500, 66.1) (4500-5500, 62.8)};
		\end{axis}
		
		\begin{axis}[bar shift=-0.15cm, hide axis]
		\addplot [draw=none, fill=gray] coordinates
		{(500-1500, 52.2) (1500-2500, 72.8) (2500-3500, 123.2) (3500-4500, 168.1) (4500-5500, 189.8)};
		\addplot [draw=none, fill=gray, fill opacity=0.4] coordinates
		{(500-1500, 13.6) (1500-2500, 40.6) (2500-3500, 75.2) (3500-4500, 85.8) (4500-5500, 87.1)};
		\end{axis}

		\begin{axis}[bar shift=0.15cm, hide axis]
		\addplot [draw=none, fill=gray!50!white] coordinates
		{(500-1500, 45.3) (1500-2500, 60.5) (2500-3500, 83.9) (3500-4500, 111.5) (4500-5500, 119.4)};
		\addplot [draw=none, fill=gray!50!white, fill opacity=0.4] coordinates
		{(500-1500, 10.5) (1500-2500, 21.6) (2500-3500, 50.0) (3500-4500, 92.3) (4500-5500, 117.6)};
		\end{axis}

		\begin{axis}[bar shift=0.45cm, legend style={/tikz/every even column/.append style={column sep=0.25cm}}]
		\addplot [draw=none, fill=black] coordinates
		{(500-1500, 0) (1500-2500, 0) (2500-3500, 0) (3500-4500, 0) (4500-5500, 0)};
		\addplot [draw=none, fill=gray] coordinates
		{(500-1500, 0) (1500-2500, 0) (2500-3500, 0) (3500-4500, 0) (4500-5500, 0)};
		\addplot [draw=none, fill=gray!50!white] coordinates
		{(500-1500, 0) (1500-2500, 0) (2500-3500, 0) (3500-4500, 0) (4500-5500, 0)};
		\addplot [draw=none, fill=orange] coordinates
		{(500-1500, 44.4) (1500-2500, 52.4) (2500-3500, 66.9) (3500-4500, 96.1) (4500-5500, 110.9)};
		\addplot [draw=none, fill=orange, fill opacity=0.4] coordinates
		{(500-1500, 12.3) (1500-2500, 16.8) (2500-3500, 34.7) (3500-4500, 67.9) (4500-5500, 107.7)};

		\legend{Uniform, Webster's, Max-pressure, Ours}
		\end{axis}
		\end{tikzpicture}
	}
	\caption{\protect\raggedright
	Performance comparison of our work with baselines on $150$ one-hour simulated demand episodes ($30$ from each of the $5$ ranges). 
    We note that the baselines are optimized for each of the test episodes 
    before they are tested on it.		
	}

    \label{fig:1h_cycle}
\end{figure}
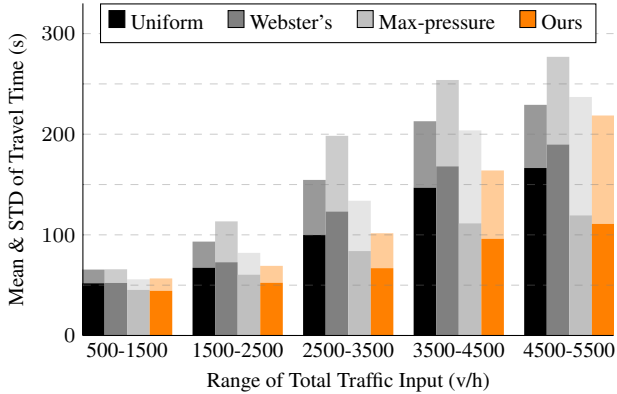

To test the performance of our agent we first evaluate on simulated traffic demand episodes that each lasts one hour.
For each of the $5$ traffic flow rate ranges as used for the evaluators during training, 
we randomly sample $30$ episodes;
this exact set of $5\cdot30$ episodes are used to test all compared algorithms. 
These demand episodes are sampled following the similar procedure to that for evaluation during training.

\begin{table}[b]
    \caption{
        Throughput ($\%$) of considered methods in Fig.~\ref{fig:1h_cycle}.
    }
    \begin{tabular}{ccccc}
		\toprule
        	\multirow{2}{*}{\shortstack{Traffic Flow Input\\(v/h)}} & \multicolumn{4}{c}{Throughput $(\%)$} \\
		\cmidrule{2-5}
			& Uniform & Webster's & Max-pressure & Ours \\
		\midrule
	 	    $\hspace{.06in}500\sim1\,500$ & $97.38$ & $97.47$ & $97.59$ & $\mathbf{97.76}$ \\
	        $1\,500\sim2\,500$ & $97.09$ & $97.52$ & $97.54$ & $\mathbf{97.95}$ \\
	        $2\,500\sim3\,500$ & $93.77$ & $95.34$ & $95.91$ & $\mathbf{97.76}$ \\
	        $3\,500\sim4\,500$ & $87.16$ & $86.65$ & $88.84$ & $\mathbf{92.88}$ \\
	        $4\,500\sim5\,500$ & $77.62$ & $74.90$ & $81.75$ & $\mathbf{86.27}$ \\
		\bottomrule
    \end{tabular}
	\label{tab:tp}
\end{table}

To ensure a fair comparison,
in each demand episode,
we use the exactly same vehicles generation time for different methods. 
Via the sampling process described above, our test set covers a very wide range of traffic scenarios and could in turn provide a more thorough evaluation.

The evaluation results are shown in Fig.~\ref{fig:1h_cycle}.
We observe that our method reaches state-of-the-art performance on all traffic flow ranges.
It is worth noting that 
for each baseline that we compare with,
we find its optimized hyperparameters for each of the $150$ test episodes;
while our agent is trained only once and a single agent is used to evaluate on all $150$ test episodes. 
This means that the overall performance of our one trained model outperforms that of the $150$ individually optimized models.
The performance improvement at about $1\,000\nicefrac{\text{v}}{\si{\hour}}$ and $5\,000\nicefrac{\text{v}}{\si{\hour}}$ is not very obvious, because in light traffic many vehicles do not have to wait in queue and in over-saturated traffic, where there is a queue in every incoming lane, the best policy is similar to scheduling the green phases cyclically. The capability of our agent to react to real time traffic situation can be fully utilized for the traffic flow ranges in the middle, where the improvement against the Max-pressure controller and the fixed-time controllers could be over $20\%$ and $40\%$.
The Webster's method performs worse than the Uniform controller due to the quick change and short duration of the test episodes, which is most of the time not the case in real traffic (Fig.~\ref{fig:real_performance}).

As mentioned, the travel time only indicates how fast the released vehicles drive through. In order to show that our agent can also benefit more drivers than baselines, we present the testing statistics for throughput in Table~\ref{tab:tp}. 
The percentage values are the ratio of the released vehicles in the total vehicle number generated. With traffic flow lower than about $3\,000\nicefrac{\text{v}}{\si{\hour}}$, all TSCs can properly release traffic input. Not $100$ percent of the generated vehicles can be released, because the test is stopped directly after one hour. Some vehicles generated at the end do not have enough time to travel through. From about $3\,000\nicefrac{\text{v}}{\si{\hour}}$ the throughput of the baselines start to drop, which means the TSC can not fully release the input traffic flow and traffic jam starts to form, while our agent can avoid traffic jams in much denser traffic.
With the increased efficiency, our agent can still guarantee equity, which is shown by the low standard deviation of vehicles travel time and the high throughput. A video of the experimental results can be found at:
\url{https://youtu.be/5-7_XpnCeKg}

\subsection{Evaluation on Real-world Traffic Demand}
To further measure the performance of our agent in more realistic traffic scenarios,
we conduct additional tests with a whole-day traffic demand of a real-world intersection of Loerracherstrasse and Wiesentalstrasse located in Freiburg, Germany. 
This intersection has different layout than
the one in Fig.~\ref{fig:intersection}.
Here each road has one forward+right turning lane with one additional short lane for protected left turn. 
So the size of the state changes to 224.
We regard the short left-turning lane, the forward+right-turning lane and the lane segment before 
the branching
as separated when we construct the state.

Since the size of the input is different from the experiments above, 
we need to train another agent. 
As we want to test the generalization capabilities of our method, 
the training traffic demand is sampled in the same way as before 
(only the maximum limit of the traffic flow is reduced to $\nicefrac{1}{2}$ to reflect the change in the intersection layout) 
The trained agent is tested on the real-world traffic demand of February 4,~2020, with typical traffic flow peaks at rush hours. The input traffic flow is in the range $[0, 1\,740]\nicefrac{\text{v}}{\si{\hour}}$. We sincerely appreciate the support of city Freiburg (www.freiburg.de/verkehr), which provides us with the traffic flow data measured with inductive-loop detectors.

\begin{figure}[t]
    \centering
    \includegraphics[width=0.49\textwidth]{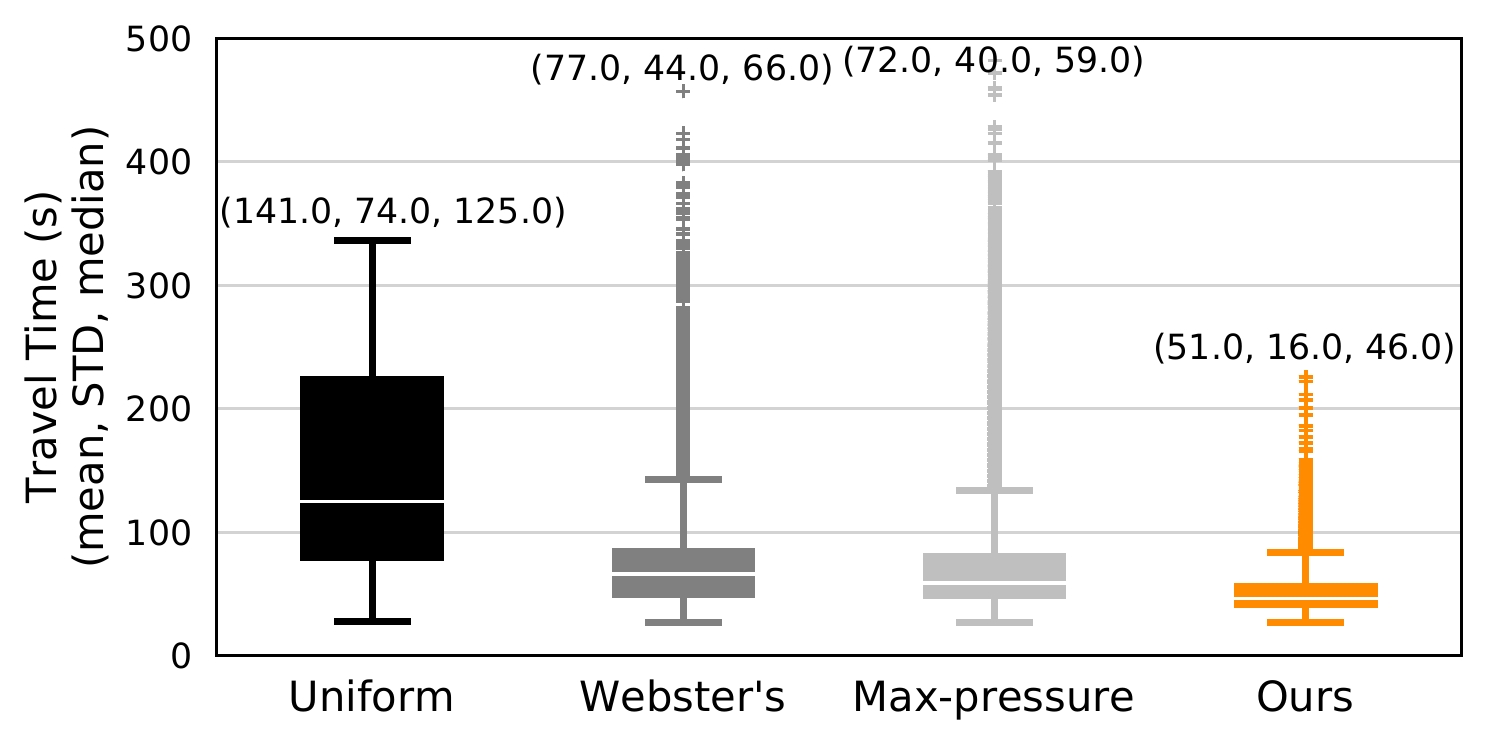}
    \caption{
    Performance comparison of our work with baseline models on a whole-day real-world traffic demand.
    }    
    \label{fig:real_performance}
\end{figure}

The results of this real-world experiment is shown in Fig.~\ref{fig:real_performance}. 
All the TSCs can properly release all vehicles, because the traffic flow is nearly zero in the night when the demand episode ends.
We can observe that our method is again outperforming all baseline methods, 
even though the baselines are firstly optimized with exactly this whole-day demand and our model is only trained on the simulated episodes with $1\,200\si{\second}$ duration. 
The substantial improvement of nearly $30\%$ on average travel time is even greater than the performance gain in the simulated evaluations. 
This validates that out proposed method has great generalization capabilities and can adapt to a wide range of traffic scenarios.

\section{Conclusion}
\label{sec:conclusion}

In this paper we presented a novel approach to learning traffic signal controllers using deep reinforcement learning. 
Our approach extends existing reward functions by a dedicated equity factor. 
We furthermore proposed a method that utilizes adaptive discounting to comply with the learning principles of deep reinforcement learning agents and to stabilize training. 
We validated the effectiveness of our approach using simulated and real-world data. Besides {substantially} outperforming  state-of-the-art methods, 
our approach is a general method that can be easily adopted to different intersection topologies.

\addtolength{\textheight}{-12cm}   



%
%

\bibliographystyle{IEEEtran}
\bibliography{yan20iros}

\end{document}